\documentclass[runningheads]{llncs}
\usepackage{graphicx}
\usepackage{amsmath}
\usepackage{amssymb}

\DeclareMathOperator*{\argmin}{arg\,min}
\begin{document}
\title{Decentralized Federated Learning Preserves Model and Data Privacy}
\titlerunning{Decentralized Federated Learning}
%
\author{Thorsten Wittkopp\inst{1}\inst{2} \and Alexander Acker\inst{1}\inst{2}}
\authorrunning{T. Wittkopp and A. Acker}
%
\institute{Technische Universität Berlin, Germany\inst{1}\\
\email{\{t.wittkopp,alexander.acker\}@tu-berlin.de}\\
Equal contribution\inst{2}
}
\maketitle              
\begin{abstract}
The increasing complexity of IT systems requires solutions, that support operations in case of failure. 
Therefore, Artificial Intelligence for System Operations (AIOps) is a field of research that is becoming increasingly focused, both in academia and industry. 
One of the major issues of this area is the lack of access to adequately labeled data, which is majorly due to legal protection regulations or industrial confidentiality.
Methods to mitigate this stir from the area of federated learning, whereby no direct access to training data is required.
Original approaches utilize a central instance to perform the model synchronization by periodical aggregation of all model parameters. 
However, there are many scenarios where trained models cannot be published since its either confidential knowledge or training data could be reconstructed from them.
Furthermore the central instance needs to be trusted and is a single point of failure.
As a solution, we propose a fully decentralized approach, which allows to share knowledge between trained models.
Neither original training data nor model parameters need to be transmitted.
The concept relies on teacher and student roles that are assigned to the models, whereby students are trained on the output of their teachers via synthetically generated input data.
We conduct a case study on log anomaly detection.
The results show that an untrained student model, trained on the teachers output reaches comparable F1-scores as the teacher.
In addition, we demonstrate that our method allows the synchronization of several models trained on different distinct training data subsets.


\end{abstract}
\section{Introduction}

IT systems are expanding rapidly to satisfy the increasing demand for a variety of applications and services in areas such as content streaming, cloud computing or distributed storage.
This entails an increasing number of interconnected devices, large networks and growing data centres to provide the required infrastructure~\cite{acker2020superiority}. 
Additionally, awareness for data privacy and confidentiality is rising especially in commercial industry.
Big- and middle-sized companies are relying on private cloud, network and storage providers to deploy and maintain according solutions.
Except for severe problems that require local access, the operation and maintenance is done remotely.
Remote access together with the growing system complexity puts extreme pressure on human operators especially when problems occur.
To maintain control and comply with defined service level agreements, operators are in need of assistance.
Therefore, monitoring solutions are combined with methods from machine learning (ML) and artificial intelligence to support the operation and maintenance of those systems - usually referred to as AIOps. 
Examples of concrete solutions are early detection of system anomalies~\cite{du2017DeepLog,nedelkoski2020self}, root cause analysis~\cite{wu2020microrca}, recommendation and automated remediation~\cite{lin2018hardware}.

The majority of ML and AI methods relies on training data.
In case of anomaly detection a common approach is to collect monitoring data such as logs, traces or metrics during normal system operation and utilize them to train models.
Representing the normal system state, these models are utilized to detect deviations from the learned representation which are labeled as anomalies.
Therefore, AIOps systems require preliminary training phases to adjust to the target environment until they can be utilized for detection.
This is known as cold start problem.
Although adjusted to customer requirements, deployed systems at different sites are very similar (e.g. private cloud solutions based on OpenStack, storage systems based on HDFS or network orchestration via ONAP).
An obvious mitigation of the cold start problem would be to use training data from existing sites to train models and fine tune them after deployment within a target customer site.
Furthermore, training data used from a variety of sites increases the holisticity of models allowing them to perform generally better.
However, sharing data or model parameters, even if indirectly related with company business cases, is usually not possible due to confidentiality or legal restrictions~\cite{shokri2015privacy}.

Federated learning as a special form of distributed learning is gaining increased attention since it allows access to a variety of locally available training data and aims to preserve data privacy~\cite{shokri2015privacy,hard2018federated}.
Utilizing this concept, we propose a method that allows different deployments of the same system to synchronize their anomaly detection models without exchanging training data or model parameters.
It does not require a central instance for model aggregation and thus, improves scalability.
We introduce a concept of student and teacher roles for models whereby student models are learning from teachers.
As input, vectors that are randomly generated within a constrained value range are used as input to both, student and teacher models.
Student models are trained on the output of teachers.
We conduct a case study based on log anomaly detection for the Hadoop File System (HDFS).
In a first experiment it is shown that our solution can mitigate the cold start problem.
A second experiment reveals that the proposed method can be utilized to holistically train distributed models.
Models that were trained by our method achieve comparable results to a model that was trained on the complete training dataset.

The rest of this paper is structured as follows. First, section~\ref{sec:rel_work} gives an overview of federated learning and applied federated learning in the field of AIOps. Second, section~\ref{sec:method} describes the our proposed method together with relevant preliminaries. Third, section~\ref{sec:eval} presents the conducted case study and experiment results. Finally, section~\ref{sec:conclusion} concludes our work.

\section{Related Work}
\label{sec:rel_work}

Federated learning is a form of distributed machine learning method. 
Thereby, model training is done locally within the environment of the data owner without sending training data to a central server.
Locality is defined within the boundaries of the data owner's private IT environment.
Initially this concept was was proposed by McMahan et al.~\cite{mcmahan2017communication}.
Instead of training data, either model weights or gradients from clients are sent to a central instance which aggregated them to a holistic model during model training.
Updated models are sent back to the clients.
Yang et al.~\cite{yang2019federated} provide a categorization, which are vertical federated learning, horizontal federated learning and federated transfer learning.
Despite preventing direct data exchange, publications revealed possibilities to restore training data from constantly transmitted weights or gradients, which violates the data privacy requirement~\cite{papernot2016semi}.
For an adversary it is possible to recover the training dataset by using a model inversion attack~\cite{fredrikson2015model} or determine whether a sample is part of the training dataset by using a membership inference attack~\cite{shokri2017membership}.
Since reconstruction of original training data is possible when observing model changes~\cite{fredrikson2015model,shokri2017membership} different privacy preserving methods are introduced.
These are focused on obfuscation of input data~\cite{hu2020sparsified} or model prediction output~\cite{bonawitz2016practical}.
Geyer et al.~\cite{geyer2017differentially} applied differential privacy preserving on client side to realize privacy protection.
Shokri and Shmatikov~\cite{shokri2015privacy} select small subsets of model parameters to be shared in order to prevent data reconstruction. 
However, model parameters or gradients still need to be shared with a central instance.
Furthermore, the requirement of a central instance is a major bottleneck for scalability~\cite{kairouz2019advances}.

Application of federated learning in the field of AIOps is mainly focused around anomaly~\cite{nguyen2019diot,liucommunication,liu2020deep} and intrusion detection~\cite{preuveneers2018chained,nguyen2020poisoning}.
Liu et al.~\cite{liu2020deep,liucommunication} propose a deep time series anomaly detection model which is trained locally on IoT devices via federated learning.
Although not directly applied on an AIOps related problem, the proposed method could be applied to perform anomaly detection on time series like CPU utilization or network traffic statistics of the device itself.
Nguyen et al.~\cite{nguyen2019diot} propose their system D{\"I}oT for detecting compromised devices in IoT networks. It utilizes an automated technique for device-type specific anomaly detection.
The unsupervised training is done via federated learning individually on each device in the IoT environment.
Preuveneers et al.~\cite{preuveneers2018chained} develop an intrusion detection system based on autoencoder models.
The model parameter exchange is coupled with a permissioned blockchain to provide integrity and prevent adversaries to alter the distributed training process.

\section{Decentralized Federated Learning}
\label{sec:method}

In this chapter we present our method for decentralized federated learning that aims at preservation of model and data privacy.
Thereby, models that were trained on individual and partly distinct training sets are synchronized.
Beside preserving data privacy, the novelty of our method is the dispensability of model parameter sharing.
We illustrate the entire process of local training and global knowledge distribution.
To realize latter, the communication process between a set of entities is described.

\subsection{Problem Definition and Preliminaries}
To apply our proposed federated learning method we assume the following setup.
Let $\Phi=\{\phi_1, \phi_2, \ldots\}$ be a set of models and $E=\{e_1, e_2, \ldots\}$ a set of environments.
We define a model deployed in a certain environment as a tuple $(\phi_i, e_j)$.
All models that perform the same task $T$ in their environment are combined into a set of workers $W_T=\{(\phi_i, e_j)\}$.
Each model $\phi_i$ has access to locally available training data but cannot directly access training data from other environments.
Furthermore, neither gradients nor model parameters can be shared outside of their environment.
Having a function $P(T,\phi_i,X_{e_j})$ that measures how well a model $\phi_i$ is performing the task $T$ on data $X_{e_j}$ from environment $e_j$, the goal is to synchronize the model training in a way that all models can perform well on data from all environments:
\begin{equation}
P(T,\phi_1,X_{e_1}) \approx P(T,\phi_1,X_{e_2}) \approx \ldots \approx P(T\phi_2,X_{e_1}) \approx P(T,\phi_2,X_{e_2}) \approx \ldots
\end{equation}
Each model $\phi_i$ is defined as a transformation function $\phi: X^{d_1} \rightarrow Y^{d_2}$ of a given input $x \in X^{d_1}$ into an output $y \in Y^{d_2}$, where $d_1$ and $d_2$ are the corresponding dimensions.
Since no original training data can be shared between environments, we define an input data range $\tilde{X}$.
It allows to draw data samples $\tilde{x} \sim \tilde{X}^{d_1}$ that are restricted to the range of the original training data but otherwise are not related to samples of the original training data set $X^{d_1}$.
Models can adopt the role of teachers $\phi^{(t)}_i$ and students $\phi^{(s)}_i$.
Student models are directly trained on the output of teachers.
We refer to a training set that is generated by a teacher as knowledge representation, formally defined as a set of tuples:
\begin{equation}
r = \{(\tilde{x}, \gamma(\phi^{(t)}_i(\tilde{x})): \tilde{x} \sim \tilde{X}^{d_1}\}.
\end{equation}
Thereby $\gamma$ is a transformation function that is applied on the teacher model output.
Student models $\phi^{(s)}_i$ are trained on the knowledge representations of teachers.
The objective is to minimize the loss between the output of teacher and student models
\begin{equation}
\argmin_{\theta^{(s)}_j}\mathcal{L}(\gamma(\phi^{(t)}_i(\tilde{x})),\phi^{(s)}_j(\tilde{x})).
\end{equation}
where $\theta^{(s)}_j$ are parameters of the student model $\phi^{(s)}_j(\tilde{x})$.

\subsection{The Concept of Teachers and Students}
\begin{figure}[htbp]
\centering
\includegraphics[width=\columnwidth]{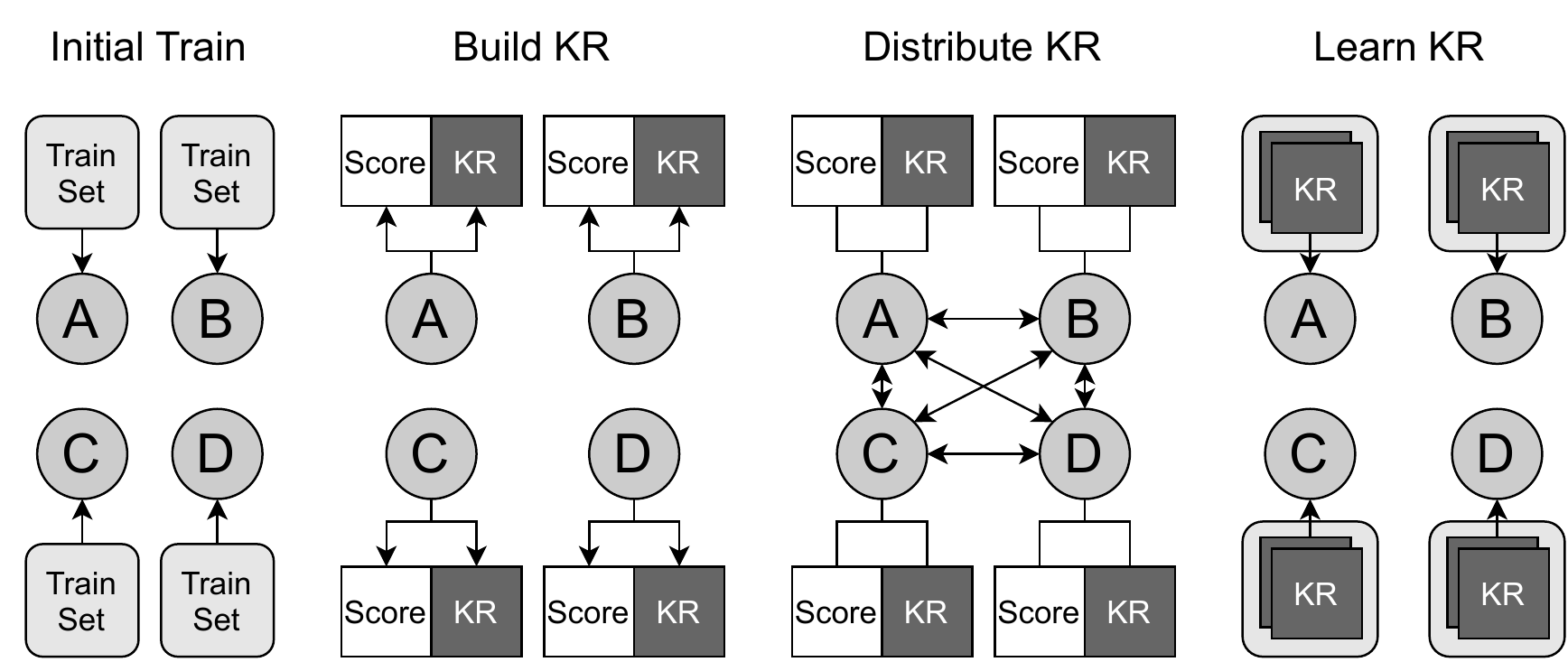}
\caption{The process of multi-cluster learning}
\label{fig:phases}
\end{figure}

For the training process itself we introduce the teacher student concept.
Every model can adopt the role of a student or teacher. 
A teacher model trains student models by providing a knowledge representation.
First, we assume a set of models, each performing the same task in their own local environment.
These models are trained on the same objective but only with the locally available training data.
To prevent the sharing of training data or model parameter between environments, models are adapting roles of teachers to train student models within other environments. 
Overall, this process is realized by four steps: (1) Initial Train, (2) adapt teacher role and build knowledge representation, (3) distribute knowledge representation and (4) adapt student role and train on teacher knowledge representation.
Figure~\ref{fig:phases} visualizes these steps the overall layout of each phase.
The example shows four environments A-D with locally available training sets.
Initially, all models are trained on the locally available training data. After that, models adapt the role of teachers and respectively generate the knowledge representations.
Thereby, auxiliary input data are generated from the value range of locally available training data.
This range must be synchronized across all environments.
Otherwise, there is no connection to original local training data that was used to train models within their environments.
Additionally, a score is calculated that reflects how well a model is performing on the task.
Next, the knowledge representations are distributed.
After receiving a knowledge representation, a model checks the attached score and compares it with its local score.
Representations with lower scores are dropped.
When receiving higher or equal scores, the models adapts the role of a student and is trained on the received knowledge representation.
Through this process each model will be retrained and updated. 
Note that the loss function used during the initial training can differ from the loss function used for knowledge representation learning.

\subsection{Loss Function}

During training of student models the objective is to directly learn the transformed outputs of a teacher for a given input.
We utilize the $tanh$ as the transformation function to restrict teacher model outputs to the range $[-1,1]$.
This restriction should reduce the output value range and thus, stabilize the training process and accelerate convergence. 
Therefore, the student needs a loss function that can minimize the loss for every element from it's own output against the output vector of the teacher.
This requires a regressive loss function.
We utilize the smooth L1 loss which calculates the absolute element-wise distance. 
If this distance falls below 1 additionally a square operation is applied on it.
It is less sensitive to outliers than the mean square error loss and prevents exploding gradients.




\section{Evaluation}
\label{sec:eval}

The evaluation of our method is done on the case study of log anomaly detection by utilizing a LSTM based neural network, called DeepLog \cite{du2017DeepLog}. DeepLog is trained on the task of predicting the next log based on a sequence of preceding log entries. However our decentralized federated learning method can be applied to any other machine learning model that is trainable via gradient descent.
We utilize the labeled HDFS data set, which is a log data set collected from the Hadoop distributed file system deployed on a cluster of 203 nodes within the Amazon EC2 platform~\cite{xu2009detecting}. 
This data set consists of 11,175,629 log lines that form a total of 570,204 labeled log sequences.
However, raw log entries are highly variant which entails a large number of prediction options.
Thus, a preprocessing is applied to reduce the number of possible prediction targets.
Log messages are transformed into templates consisting of a constant and variable part.
The constant templates are used as discrete input and prediction target.
The task of template parsing is executed by Drain~\cite{he2017drain}.
We refer to the set of templates as $\mathbb{T}$.

\subsection{Auxiliary Sample Generation}\label{seq:sample_gen}
As described, student models are trained on auxiliary samples together with teacher model outputs.
These samples are drawn from a restricted range but otherwise independent from the training data that was used to train the teacher.
In the conducted use case study of log anomaly detection auxiliary samples are randomly generated as follows.
Having $\mathbb{T}$ as the set of unique templates, an auxiliary sample is defined as $\tilde{x}=(t_i \sim \mathbb{T} : i=1,2,\ldots,w+1)$, where $w$ is the input sequence length.
Template $t_{w+1}$ will be used as the prediction target.
Note that templates are randomly sampled from the unique template set $\mathbb{T}$.
Thus, auxiliary input samples for DeepLog are random template sequences.



\subsection{Experiment 1: Training of an Untrained Model}\label{sec:exp_1}
In our first experiment we investigate the ability of the proposed method to mitigate the cold-start problem.
Therefore, a DeepLog model is trained on the HDFS data.
Out of the 570,204 labeled log sequences of the HDFS data set, we use the first 4,855 normal sequences as training and the remaining 565,349 as test set.
The test set contains 16,838 anomalies and 553,366 normal sequences.
The two hyper-parameters of DeepLog were set as follow: $w=10$ and $k=9$, where $w$ is the window size that determines the input sequence length and is required to generate auxiliary samples.
Further we use cross-entropy loss to learn a distribution when predicting possible next log lines.
The teacher model uses a batch size of 16 and we trained it over 20 epochs on all 4,855 normal log sequences.
The therewith trained DeepLog model is utilized as the teacher while a completely untrained model with the same architecture and parametrization adopts the role of a student. 
The teacher performs following transformation: $\phi: X^{w \times |\mathbb{T}|} \rightarrow Y^{|\mathbb{T}| \times \mathbb{R}^{[-1,1]}}$ to generate the knowledge representation.
It takes a sequence with $w$ one-hot encoded templates and outputs a $tanh$ transformed probability distribution over all $w$ templates.
Different amounts of knowledge representation tuples are tested: $\{10, 50, 100, 500, 1000, 5000\}$.
The student model uses a batch size of 16 and we trained it over 10 epochs for every knowledge representation size.
\begin{figure}[htbp]
\centering
\includegraphics[width=8cm]{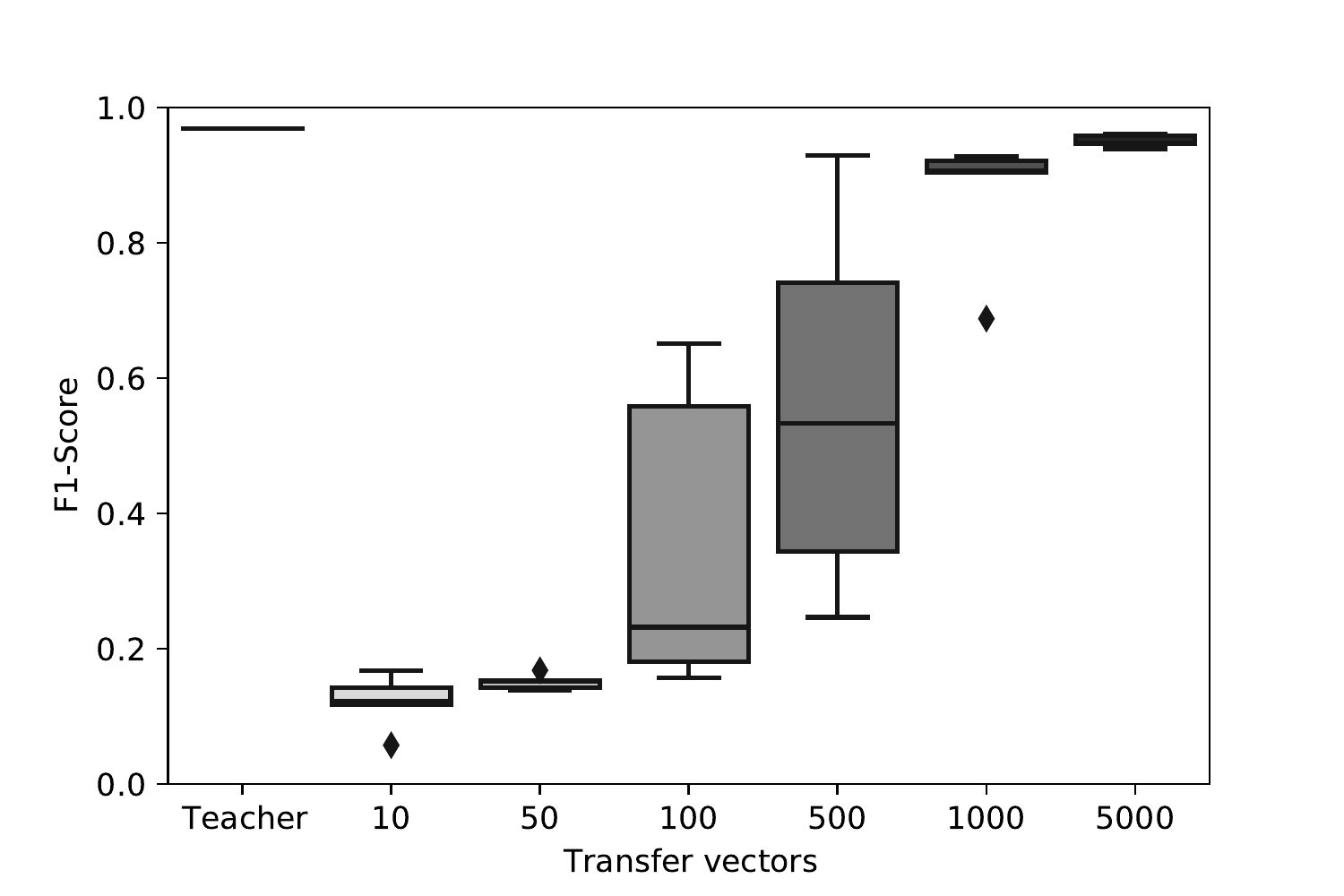}
\caption{Comparison between F1-scores of teacher model and student models that were trained on different knowledge representation sizes.}
\label{fig:res_transfer}
\end{figure}
Auxiliary input samples are generated as described in section~\ref{seq:sample_gen}.
Due to this randomness of sample generation, we repeat the experiment five times. 
Figure~\ref{fig:res_transfer} shows the results as a boxplot, which illustrated the the F1-scores for the teacher and students. 
Bottom and top whiskers reflect the minimum and maximum non-outlier values. 
The line in the middle of the box represents the median. 
The box boundaries are the first quartile and third quartile of the value distribution.
The most left bar shows the F1-score for the teacher model, which is 0.965. This bar is a flat line, because it represents a single value.
Remaining bars are visualizing the F1-score for each knowledge representation size, from 10 to 5,000.
First, it can be observed that a knowledge representation size of at least 100 is required to have a decent F1-score on the student model.
As expected, the score of the student increases with the number of used knowledge representation tuples.
Due to the random sampling F1-scored of student models trained with 100 and 500 knowledge representation tuples underlie high uncertainty.
The 0.95 confidence interval ranges from 0.153 to 0.558 for size 100 and from 0.313 to 0.805 for size 500.
The student model's F1-score becomes increasingly stable with higher knowledge representation sizes and reaches 0.961 within the 0.95 confidence interval of $[0.943,0.958]$ for size 5000.
Compared to the teacher model's F1-score of 0.965 we conclude that our method can be utilized to mitigate the cold-start problem by training an untrained model via knowledge representations of trained models.



\subsection{Experiment 2: Federated Learning}\label{sec:exp_2}
In this experiment we investigate how multiple DeepLog models behave as teachers and students.
It allows to train distributed models on locally available training data and subsequently synchronize the knowledge of models.
Therefore, we simulate 8 distributed HDFS systems by creating a set of unique sequences. 
Out of the 570,204 labeled log sequences of the HDFS data set, we again use the first 4,855 normal sequences to generate unique sequences of size $w$. 
This results in 4,092 unique training samples. 
These 4,092 training samples were evenly and randomly split into 8 sets, hence every set contains 511 training samples.
Therefore, 8 DeepLog models are respectively trained. 
The two hyper-parameters of each DeepLog were set as follow: $w=10$ and $k=9$.
To evaluate the model performance on predicting potentially unseen samples, remaining 565,349 sequences are used as a joint test set.


We initially trained the 8 DeepLog models over 10 epochs with a batch size of 16 and cross-entropy-loss as a loss function. 
The therewith trained DeepLog model are utilized as teacher models for each other.
Hence, all 8 nodes are also students and where trained with the knowledge representation from the teachers with a batch size of 1 over 5 epochs.
Again, we test different amounts of knowledge representation per model: $\{10, 50, 100, 500, 1000\}$.
Note that a student model is trained on the representations of all teacher models that have a higher or equal score than itself.

\begin{figure}[htbp]
\centering
\includegraphics[width=10cm]{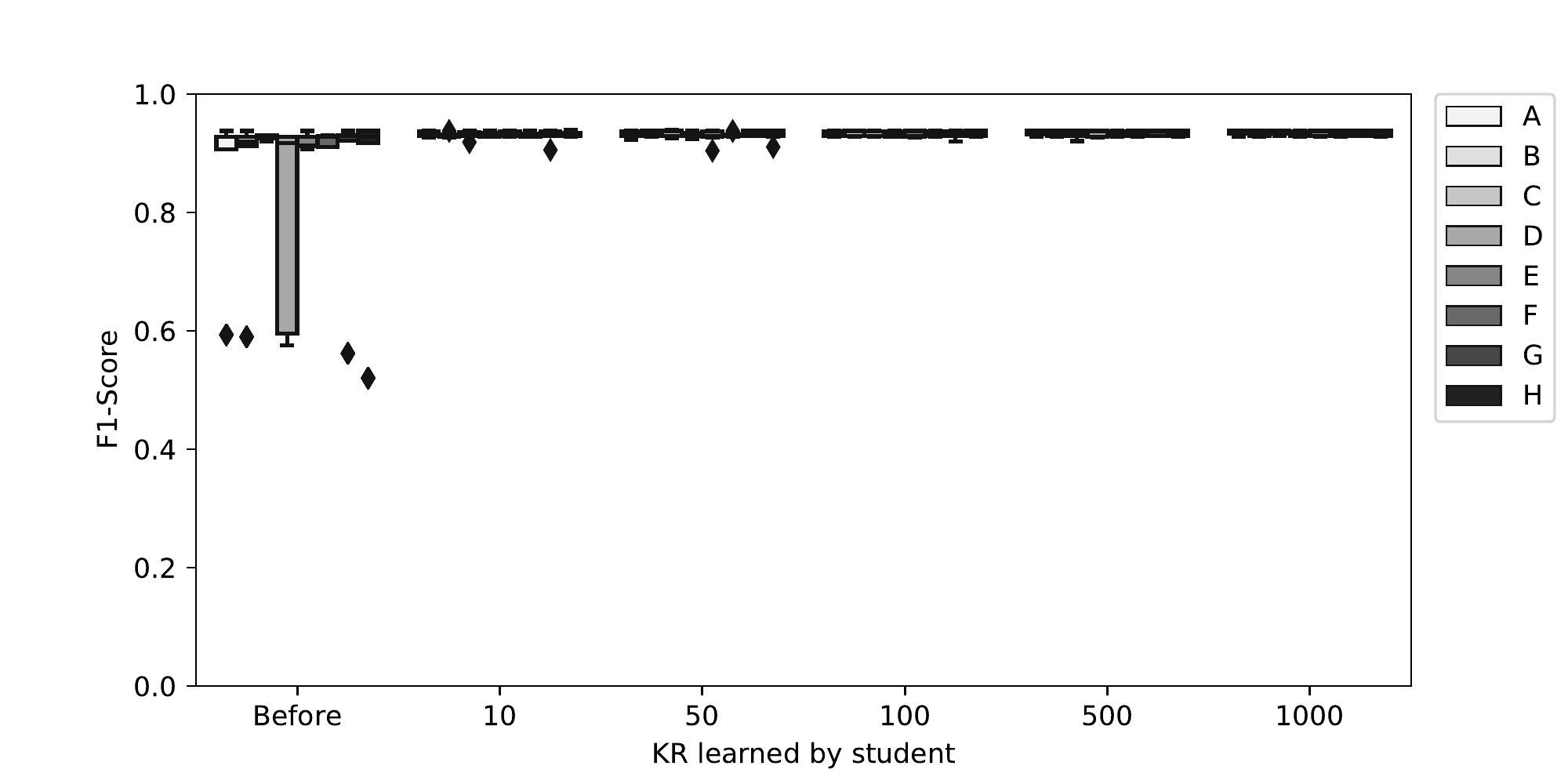}
\caption{Comparison of F1-scores of each node before and after the teacher student process with different knowledge representation sizes.}
\label{fig:exp2_res_multitransfer}
\end{figure}

Auxiliary input samples are generated as described in section~\ref{seq:sample_gen}.
Due to this random generation of auxiliary samples, we repeat the experiment 7 times. 
Figure~\ref{fig:exp2_res_multitransfer} shows the results as a boxplot, which illustrates the F1-scores of all 8 models (marked as A-H) for different amounts of knowledge representation tuples over 7 experiment executions. 
The properties of the boxplot are the same as described in~\ref{sec:exp_1}.
The most left bars in the category \textit{Before} for each model after initially trained on the locally available unique sequence set.
No federated learning is applied here.
It can be seen that their F1-scores ranging from 0.520 to 0.938.
The 0.95 confidence interval for all 8 nodes in this section ranges from 0.875 to 0.901.


\begin{figure}[htbp]
\centering
\includegraphics[width=10cm]{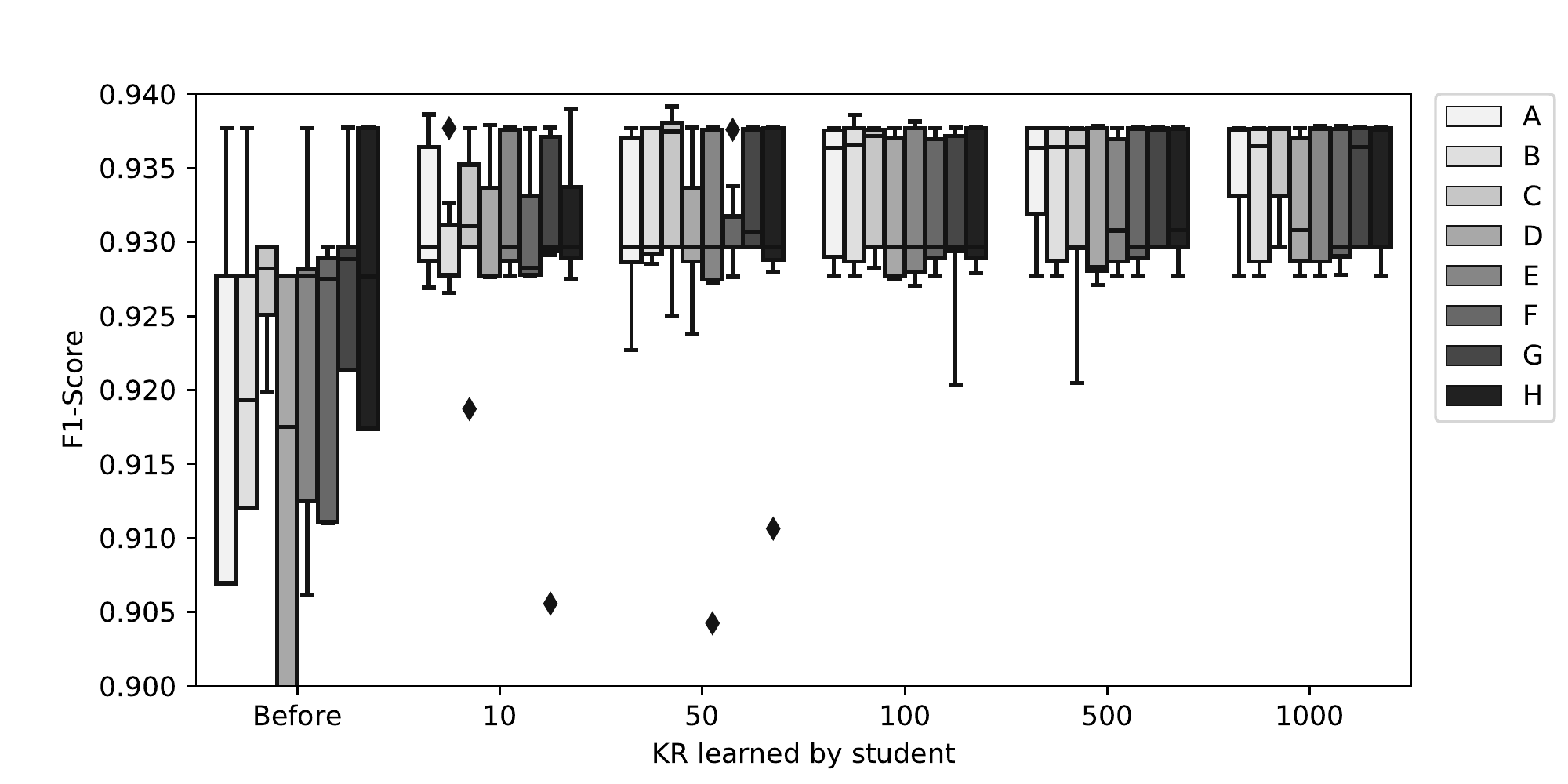}
\caption{Comparison of F1-scores between 0.90 and 0.94 of each node before and after the teacher student process with different knowledge representation sizes.}
\label{fig:exp2_res_multitransfer_zoom}
\end{figure}

The figure \ref{fig:exp2_res_multitransfer_zoom} shows a zoomed-in version of the same boxplot on the F1-score range of between 0.90 and 0.94. 
As expected, an increasing amount of knowledge representations leads to overall higher F1-scores.
Furthermore, it is visible that already 10 knowledge representations are improving the F1-scores significantly.
After 10 knowledge representation the lowest F1-score is 0.906 and the 0.95 confidence interval for this section ranges from 0.923 to 0.933. 
In the category of 50 knowledge representations, the number of existing outliers is the same, but the 0.95 confidence interval ranges from 0.930 to 0.933, which is an improvement compared to 10 knowledge representations. 
Also the highest F1-score of 0.939 could be observed in this category. 
The lowest deviations of the F1-score for all nodes occur at 1000 knowledge representations with a 0.95 confidence interval from 0.933 to 0.935. 
In this category the highest F1-score is 0.938 and the lowest 0.928. 
Compared to the 0.95 confidence interval of 0.875 to 0.901 before the teacher student process, we observe an improvement in every category.

This experiment indicates, that it is able to train different models with the same configuration in a distributed system.
The method preserves data privacy by using auxiliary samples to transfer knowledge between models.
A central instance for synchronization of the training process is not required.
Neither model parameters nor gradients need to be transferred.



\section{Conclusion}
\label{sec:conclusion}
In this work we proposed a federated learning solution for synchronizing distributed models trained on locally available data.
Out method does not required a sharing of original training data or model parameter. 
Training is done via assignment of teacher and student roles to existing models.
Students are trained on the output of teachers via auxiliary samples and respective outputs, referred to as knowledge representations.
We evaluated our approach in a case study of log anomaly detection. 
DeepLog models were trained on distinct and unique log sequences from the HDFS data set.
After that the teacher and student roles were applied to the models in order to test the ability of synchronizing them.
In our first experiment we could show that this approach can mitigate the cold start problem. 
For this experiment we setup a trained teacher DeepLog model and an untrained DeepLog model as a student. 
We investigated how well they student adapts the teacher with different amounts of knowledge representations. 
After applying the proposed method, the student model achieved a comparable F1-score of 0.96 while the teacher achieved 0.97.
In the second experiment, we demonstrated that our method allows the synchronization of several models trained on different distinct training data sets through the proposed decentralized federated learning process. 
Therefore, we split the training set into 8 equal and unique log sequence subsets and distributed these among 8 DeepLog models. 
With this training data all nodes could perform an initial training step before they entered the role of teachers and students. 
Even with distributing small amount of 10 knowledge representations all nodes could improve to a F1-score 0.95 confidence interval between 0.923 and 0.933 compared to their initial trained models, which reached a 0.95 confidence interval of 0.875 to 0.901. 
After 1,000 the models archive a 0.95 confidence interval of between 0.933 and 0.935.

For future work we plan to investigate more datasets in order to verify the general applicability of our approach.
Furthermore, generating random sequences from a relatively small set of discrete elements represents a comparably limited search space for auxiliary sample generation.
We expect the knowledge transfer to be harder within larger discrete sets or even within continuous space.
Another goal is to research methods and heuristics to stabilize and accelerate the process of knowledge transfer with increasingly complex auxiliary samples. 

\bibliographystyle{IEEEtran}
\bibliography{main}

\begin{thebibliography}{10}
\providecommand{\url}[1]{#1}
\csname url@samestyle\endcsname
\providecommand{\newblock}{\relax}
\providecommand{\bibinfo}[2]{#2}
\providecommand{\BIBentrySTDinterwordspacing}{\spaceskip=0pt\relax}
\providecommand{\BIBentryALTinterwordstretchfactor}{4}
\providecommand{\BIBentryALTinterwordspacing}{\spaceskip=\fontdimen2\font plus
\BIBentryALTinterwordstretchfactor\fontdimen3\font minus
  \fontdimen4\font\relax}
\providecommand{\BIBforeignlanguage}[2]{{%
\expandafter\ifx\csname l@#1\endcsname\relax
\typeout{** WARNING: IEEEtran.bst: No hyphenation pattern has been}%
\typeout{** loaded for the language `#1'. Using the pattern for}%
\typeout{** the default language instead.}%
\else
\language=\csname l@#1\endcsname
\fi
#2}}
\providecommand{\BIBdecl}{\relax}
\BIBdecl

\bibitem{acker2020superiority}
A.~Acker, T.~Wittkopp, S.~Nedelkoski, J.~Bogatinovski, and O.~Kao,
  ``Superiority of simplicity: A lightweight model for network device workload
  prediction,'' \emph{arXiv preprint arXiv:2007.03568}, 2020.

\bibitem{du2017DeepLog}
M.~Du, F.~Li, G.~Zheng, and V.~Srikumar, ``Deeplog: Anomaly detection and
  diagnosis from system logs through deep learning,'' in \emph{Proceedings of
  the 2017 ACM SIGSAC Conference on Computer and Communications Security},
  2017, pp. 1285--1298.

\bibitem{nedelkoski2020self}
S.~Nedelkoski, J.~Bogatinovski, A.~Acker, J.~Cardoso, and O.~Kao,
  ``Self-supervised log parsing,'' \emph{arXiv preprint arXiv:2003.07905},
  2020.

\bibitem{wu2020microrca}
L.~Wu, J.~Tordsson, E.~Elmroth, and O.~Kao, ``Microrca: Root cause localization
  of performance issues in microservices,'' in \emph{IEEE/IFIP Network
  Operations and Management Symposium (NOMS)}, 2020.

\bibitem{lin2018hardware}
F.~Lin, M.~Beadon, H.~D. Dixit, G.~Vunnam, A.~Desai, and S.~Sankar, ``Hardware
  remediation at scale,'' in \emph{2018 48th Annual IEEE/IFIP International
  Conference on Dependable Systems and Networks Workshops (DSN-W)}.\hskip 1em
  plus 0.5em minus 0.4em\relax IEEE, 2018, pp. 14--17.

\bibitem{shokri2015privacy}
R.~Shokri and V.~Shmatikov, ``Privacy-preserving deep learning,'' in
  \emph{Proceedings of the 22nd ACM SIGSAC conference on computer and
  communications security}, 2015, pp. 1310--1321.

\bibitem{hard2018federated}
A.~Hard, K.~Rao, R.~Mathews, S.~Ramaswamy, F.~Beaufays, S.~Augenstein,
  H.~Eichner, C.~Kiddon, and D.~Ramage, ``Federated learning for mobile
  keyboard prediction,'' \emph{arXiv preprint arXiv:1811.03604}, 2018.

\bibitem{mcmahan2017communication}
B.~McMahan, E.~Moore, D.~Ramage, S.~Hampson, and B.~A. y~Arcas,
  ``Communication-efficient learning of deep networks from decentralized
  data,'' in \emph{Artificial Intelligence and Statistics}, 2017, pp.
  1273--1282.

\bibitem{yang2019federated}
Q.~Yang, Y.~Liu, T.~Chen, and Y.~Tong, ``Federated machine learning: Concept
  and applications,'' \emph{ACM Transactions on Intelligent Systems and
  Technology (TIST)}, vol.~10, no.~2, pp. 1--19, 2019.

\bibitem{papernot2016semi}
N.~Papernot, M.~Abadi, U.~Erlingsson, I.~Goodfellow, and K.~Talwar,
  ``Semi-supervised knowledge transfer for deep learning from private training
  data,'' \emph{arXiv preprint arXiv:1610.05755}, 2016.

\bibitem{fredrikson2015model}
M.~Fredrikson, S.~Jha, and T.~Ristenpart, ``Model inversion attacks that
  exploit confidence information and basic countermeasures,'' in
  \emph{Proceedings of the 22nd ACM SIGSAC Conference on Computer and
  Communications Security}, 2015, pp. 1322--1333.

\bibitem{shokri2017membership}
R.~Shokri, M.~Stronati, C.~Song, and V.~Shmatikov, ``Membership inference
  attacks against machine learning models,'' in \emph{2017 IEEE Symposium on
  Security and Privacy (SP)}.\hskip 1em plus 0.5em minus 0.4em\relax IEEE,
  2017, pp. 3--18.

\bibitem{hu2020sparsified}
R.~Hu, Y.~Gong, and Y.~Guo, ``Sparsified privacy-masking for
  communication-efficient and privacy-preserving federated learning,''
  \emph{arXiv preprint arXiv:2008.01558}, 2020.

\bibitem{bonawitz2016practical}
K.~Bonawitz, V.~Ivanov, B.~Kreuter, A.~Marcedone, H.~B. McMahan, S.~Patel,
  D.~Ramage, A.~Segal, and K.~Seth, ``Practical secure aggregation for
  federated learning on user-held data,'' \emph{arXiv preprint
  arXiv:1611.04482}, 2016.

\bibitem{geyer2017differentially}
R.~C. Geyer, T.~Klein, and M.~Nabi, ``Differentially private federated
  learning: A client level perspective,'' \emph{arXiv preprint
  arXiv:1712.07557}, 2017.

\bibitem{kairouz2019advances}
P.~Kairouz, H.~B. McMahan, B.~Avent, A.~Bellet, M.~Bennis, A.~N. Bhagoji,
  K.~Bonawitz, Z.~Charles, G.~Cormode, R.~Cummings \emph{et~al.}, ``Advances
  and open problems in federated learning,'' \emph{arXiv preprint
  arXiv:1912.04977}, 2019.

\bibitem{nguyen2019diot}
T.~D. Nguyen, S.~Marchal, M.~Miettinen, H.~Fereidooni, N.~Asokan, and A.-R.
  Sadeghi, ``D{\"i}ot: A federated self-learning anomaly detection system for
  iot,'' in \emph{2019 IEEE 39th International Conference on Distributed
  Computing Systems (ICDCS)}.\hskip 1em plus 0.5em minus 0.4em\relax IEEE,
  2019, pp. 756--767.

\bibitem{liucommunication}
Y.~Liu, N.~Kumar, Z.~Xiong, W.~Y.~B. Lim, J.~Kang, and D.~Niyato,
  ``Communication-efficient federated learning for anomaly detection in
  industrial internet of things.''

\bibitem{liu2020deep}
Y.~Liu, S.~Garg, J.~Nie, Y.~Zhang, Z.~Xiong, J.~Kang, and M.~S. Hossain, ``Deep
  anomaly detection for time-series data in industrial iot: A
  communication-efficient on-device federated learning approach,'' \emph{IEEE
  Internet of Things Journal}, 2020.

\bibitem{preuveneers2018chained}
D.~Preuveneers, V.~Rimmer, I.~Tsingenopoulos, J.~Spooren, W.~Joosen, and
  E.~Ilie-Zudor, ``Chained anomaly detection models for federated learning: An
  intrusion detection case study,'' \emph{Applied Sciences}, vol.~8, no.~12, p.
  2663, 2018.

\bibitem{nguyen2020poisoning}
T.~D. Nguyen, P.~Rieger, M.~Miettinen, and A.-R. Sadeghi, ``Poisoning attacks
  on federated learning-based iot intrusion detection system,'' 2020.

\bibitem{xu2009detecting}
W.~Xu, L.~Huang, A.~Fox, D.~Patterson, and M.~I. Jordan, ``Detecting
  large-scale system problems by mining console logs,'' in \emph{Proceedings of
  the ACM SIGOPS 22nd symposium on Operating systems principles}, 2009, pp.
  117--132.

\bibitem{he2017drain}
P.~He, J.~Zhu, Z.~Zheng, and M.~R. Lyu, ``Drain: An online log parsing approach
  with fixed depth tree,'' in \emph{2017 IEEE International Conference on Web
  Services (ICWS)}.\hskip 1em plus 0.5em minus 0.4em\relax IEEE, 2017, pp.
  33--40.

\end{thebibliography}

\end{document}